\documentclass[10pt,twocolumn,letterpaper]{article}

\usepackage{iccv}
\usepackage{times}
\usepackage{epsfig}
\usepackage{graphicx}
\usepackage{amsmath}
\usepackage{amssymb}
\usepackage{multirow}
\usepackage[flushleft]{threeparttable}
\usepackage{subfig}
\usepackage{placeins}
\usepackage{enumitem}
\usepackage{color}
\usepackage{svg}
\usepackage[pagebackref=true,breaklinks=true,letterpaper=true,colorlinks,bookmarks=false]{hyperref}
\usepackage{cleveref}
\usepackage{booktabs}
\usepackage{balance}
\usepackage{caption}

\iccvfinalcopy 

\begin{document}

\title{VizWiz Dataset Browser: A Tool for Visualizing Machine Learning Datasets}


\author{Nilavra Bhattacharya$^*$, Danna Gurari \\
\noindent
{\small University of Texas at Austin}\\
{\tt\small $^*$nilavra@ieee.org}
}

\twocolumn[{%
\renewcommand\twocolumn[1][]{#1}%
\maketitle
\begin{center}
\centering
\vspace{-2em}
\fbox{\includegraphics[clip=true, trim=0 0 0 0, width=\textwidth]{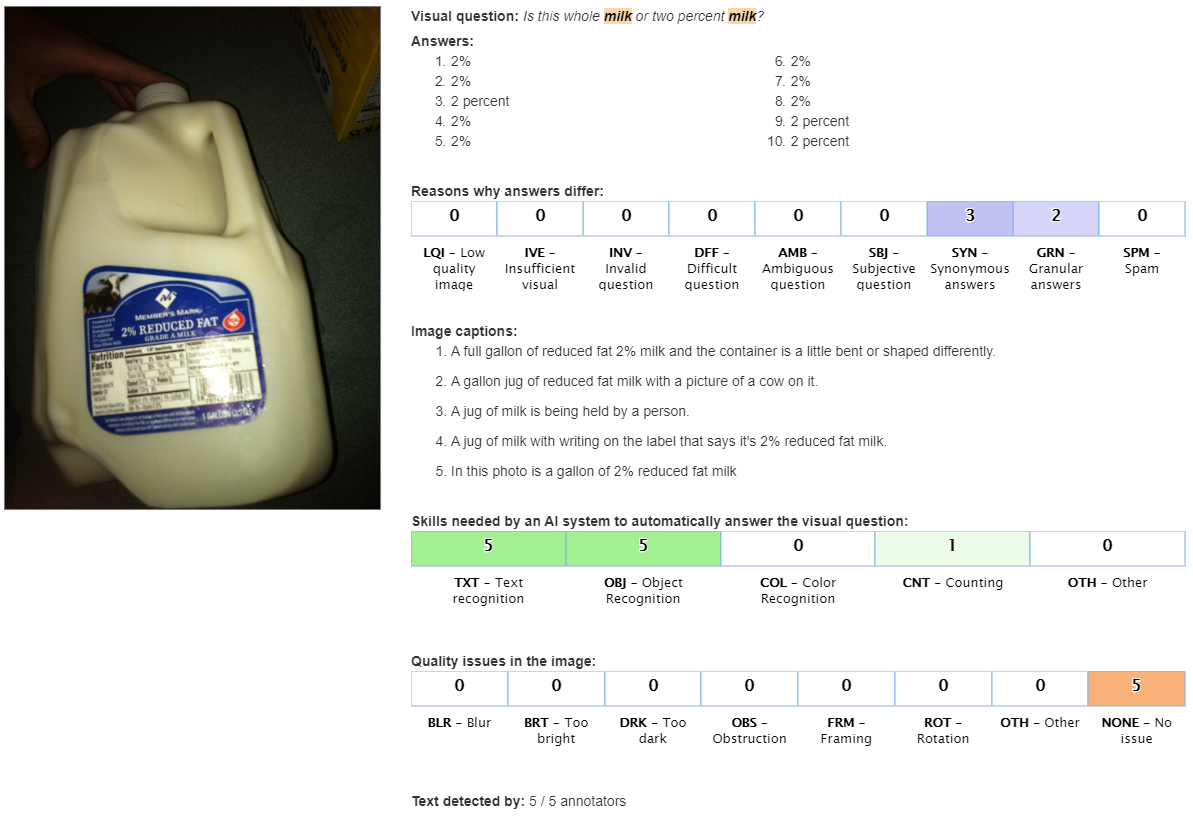}} 
\captionof{figure}{
Screenshot of our visualization tool showing an example of a visual-question, alongside different types of annotations that can be searched for.
This example was obtained by searching for the word ``milk'' within the question text.
}
\label{fig:viz_section}
\end{center}%
}]

\vspace{-1em}
\begin{abstract} 
\vspace{-1em}
We present a visualization tool to exhaustively search and browse through a set of large-scale machine learning datasets.
Built on the top of the VizWiz dataset, our dataset browser tool has the potential to support and enable a variety of qualitative and quantitative research, and open new directions for visualizing and researching with multimodal information. 
The tool is publicly available at
\url{https://vizwiz.org/browse}.
\end{abstract}

\vspace{-1em}
\section{Introduction}
\label{sec:intro}
A major challenge of working with large-scale machine learning datasets is the difficulty of exploratory data analysis \cite{tukey1977exploratory}. 
Researchers often want to immerse themselves in the data. 
For datasets containing thousands of images and annotations, there is no straightforward way to do this. 
Visualization efforts involve writing very specific programs or scripts to generate plots, such as bar charts containing counts of different categories, or sunburst diagrams showing relative proportions of different annotations. 
However, these visualization attempts can produce aggregated results only, thereby hiding interesting examples. 
Even for data cleaning and quality control purposes, manually going through each individual image and its annotation is tedious, and is prone to human error.

To overcome these challenges, we developed the VizWiz dataset browser. 
The VizWiz dataset originates from people who are blind, who used mobile phones to snap photos and record questions about them (\eg, ``what type of beverage is in this bottle?'' or ``has the milk expired?''), and contains images paired with questions about the image \cite{bigham2010VizWiznearlyrealtime}. 
Subsequent research has generated a variety of annotations on top the VizWiz dataset. 
These include: 
ten crowdsourced answers to each visual question \cite{gurari2018VizWizGrandChallenge},
reasons explaining why the ten answers can differ, if they do \cite{bhattacharya2019does}; 
captions for describing the images to users with visual impairments; 
the multitude of skills needed by an AI system to automatically answer the visual question; 
quality issues present in the images (since they were captured by users who could not see the photo they were capturing), 
and
whether text is present in the image. 
As more and more annotations were being collected, we felt the need to view all these different kinds of rich data in a single platform, in order to get a holistic view of the information contained within these datasets.

\section{Design and Implementation}
\label{sec:design_imple}

The VizWiz Dataset Browser is a single-page web-application built on the Linux - Apache - MariaDB - PHP (LAMP) stack. 
It supports searching for textual annotations, and filtering for categorical annotations. 
The main purpose of the tool is to view images, and search for those images using the `meta-data' provided by the annotations.
To scale effortlessly with an increasing variety of annotations, we decided to keep the search functionalities on the left side of the screen, in its own independently scrollable section. 
By not opting for a horizontal layout of the search and filter options, we can display more dynamic information above the fold.
Similar design choices are employed by popular eCommerce websites which display numerous filters on their search-results page \cite{url_amazon, url_ebay, url_walmart}.

\subsection{Visualization Section}
\label{sec:viz_section}
Figure 1 shows a screenshot of the main information visualization area.
The image and the \textbf{textual annotations}: (a) question, (b) ten answers, and (c) five captions, are displayed in their natural form, while 
the \textbf{categorical annotations}: (d) answer-difference reasons, (e) skills, and (f) quality issues, are displayed as one-dimensional heatmaps, based on how many crowdworkers (out of 5) selected a categorical label.

\subsection{Summary of Results}
\label{sec:results_summary}
\begin{figure}[!h]
\centering
\fbox{\includegraphics[width=\linewidth]{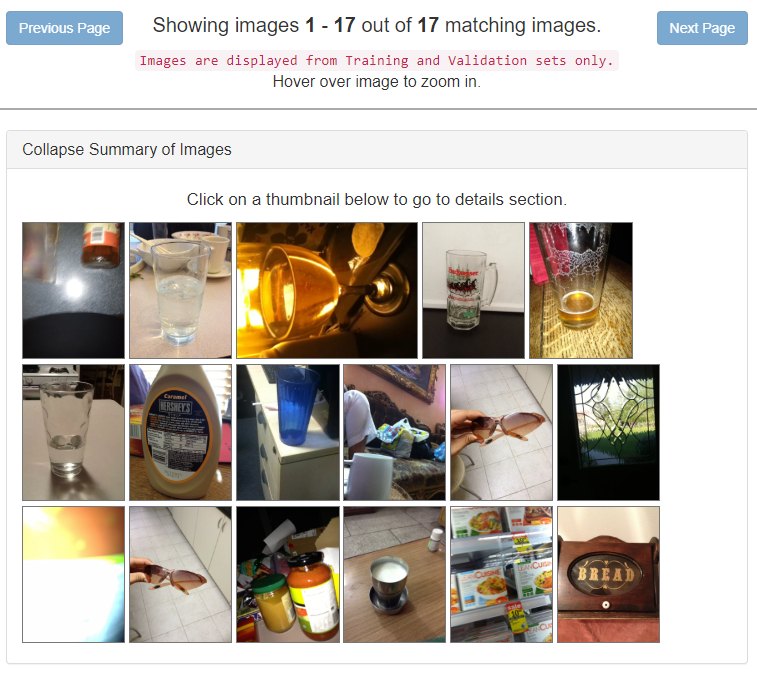}}
\caption{
The summary section shows an overview of the different images returned for the search or filter query.
Clicking a thumbnail image lets the user view the details of the image, as in Figure~\ref{fig:viz_section}.
This example was obtained by searching for the word ``glass'' in the question.
}
\label{fig:results_summary}
\end{figure}

The top portion of the visualization section shows a summary of the search results. 
This includes the number of total images found for the current search and/or filter query, and the range of images shown on the current page.
To support minimal page loading times, we decided to show a maximum of 50 images per page.
Users can choose to view the thumbnails of all the images displayed on the current page (as shown in Figure~\ref{fig:results_summary}) by clicking on `Expand Summary of Images'.
Clicking on a thumbnail image within the `Summary of Images' section will take the user to the details-section of the image.

\subsection{Searching for Images by Textual Annotations}
\label{sec:viz_search_text}
\begin{figure}[!h]
\centering
\fbox{\includegraphics[width=0.8\linewidth]{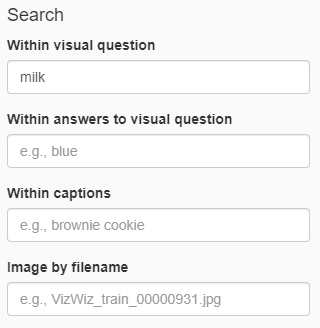}}
\caption{Different ways to search for images using textual annotations. 
Users can search for words and phrases within the question, the ten answers, and the five captions.
}
\label{fig:search_text}
\end{figure}

Text searching capabilities are present for searching for words and phrases within the visual question, the ten answers, and the five crowdsourced captions.
Full-text searching is powered by MariaDB relational database\footnote{\url{https://mariadb.com/kb/en/library/full-text-index-overview}}. 
Additionally, users can search for an image using its specific filename.
These search capabilities are shown in Figure~\ref{fig:search_text}.

\subsection{Filtering Images by Categorical Annotations}
\label{sec:viz_filter_categorical}
\begin{figure}[!h]
\centering
\fbox{\includegraphics[width=0.8\linewidth]{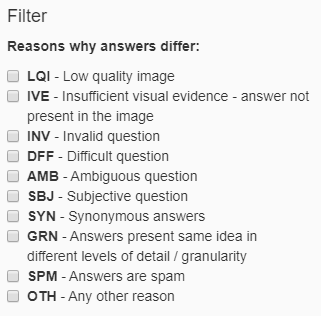}}
\caption{
Filtering for images using categorical annotations. The screenshot shows the labels for the answer-difference dataset \cite{bhattacharya2019does}.
}
\label{fig:filter_categorical}
\end{figure}
The visualization tool can be used to filter images based on the different types of categorical annotations available: (a) answer-difference reasons, (b) skills, and (c) quality issues.
This functionality proves to be useful when we want to explore relationships between the different datasets.
For example, selecting DFF (Difficult Question) as an answer-difference reason, and ROT (image needs to be rotated) as an image-quality issue, we can view the specific cases where the visual questions are difficult to answer because the images need to be rotated.
The filtering capabilities for the answer-difference reasons are shown in Figure~\ref{fig:filter_categorical}.

\subsection{Ordering of Search Results}
\label{sec:viz_order_by}
\begin{figure}[!h]
\centering
\fbox{\includegraphics[width=0.8\linewidth]{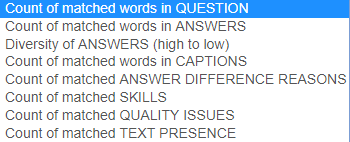}}
\caption{
Various options for ordering the search results.
}
\label{fig:order_by}
\end{figure}

The search results can be ordered (sorted) using the options shown in Figure~\ref{fig:order_by}.
When searching for textual annotations (words or phrases in the question, answers, or captions), the result are sorted in decreasing order of the number of matched words in the annotation.
`Diversity of answers' orders the results based on how different the ten answers are, using the Shannon Entropy of the ten answers.
For categorical annotations (answer-difference reasons, skills, quality issues, text-presence), the results are ranked based on how many crowdworkers (out of five) annotated the images using the chosen categorical labels.

\subsection{Toggling Display of Annotations}
\label{sec:viz_view_ann}
\begin{figure}[!h]
\centering
\fbox{\includegraphics[width=0.8\linewidth]{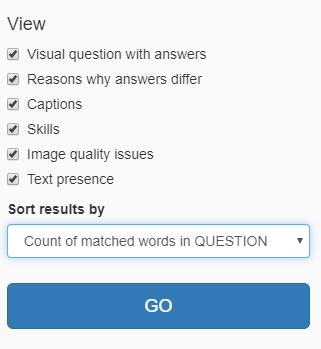}}
\caption{Options to hide or show different datasets.}
\label{fig:view_annotations}
\end{figure}
Viewing all the different annotations at once can be overwhelming.
Often, the user may want to selectively view certain annotations (\eg, for taking screenshots).
For this purpose, the `View' section, as shown in Figure~\ref{fig:view_annotations}, can be used to hide or show the different datasets as desired.

\section{Conclusion}
\label{sec:conclusion}
In summary, the VizWiz Dataset Browser can prove to be a useful tool to search, filter, and visualize multiple large datasets.
It is already being used to aid a variety of ongoing research efforts in the domains of computer vision, accessibility, and human-computer interaction.
We are hopeful that future researchers who choose to work with the VizWiz dataset will find the tool useful for answering interesting research questions.

\noindent
\section*{Acknowledgements}
\noindent
We thank the crowdworkers for providing the annotations. 
We thank Kenneth R. Fleischmann, Meredith Morris, Ed Cutrell, and Abigale Stangl for their valuable feedback about this tool and paper.  
This work is supported in part by funding from the National Science Foundation (IIS-1755593) and Microsoft.

{\small
\bibliographystyle{ieee}
\bibliography{manuscript}
}

\end{document}